\documentclass{article}
\usepackage{ijcai25}

\usepackage{times}
\usepackage{soul}
\usepackage{url}
\usepackage[hidelinks]{hyperref}
\usepackage[utf8]{inputenc}
\usepackage[small]{caption}
\usepackage{graphicx}
\usepackage{amsmath}
\usepackage{amsthm}
\usepackage{booktabs}
\usepackage{algorithm}
\usepackage{algorithmic}
\usepackage[switch]{lineno}

\usepackage{tikz}
\usetikzlibrary{decorations.pathreplacing}
\usetikzlibrary{arrows.meta}
\usetikzlibrary[shapes.geometric]
\usepackage{amsfonts}
\usepackage{amssymb}
\usepackage{stmaryrd}

\usepackage{enumitem}
\usepackage{booktabs}
\usepackage{subcaption}
\usepackage{enumitem}
\usepackage{fvextra}
\usepackage{csquotes}
\usepackage{xcolor}

\definecolor{LightGray}{RGB}{230, 230, 230}

\usepackage{newfloat}
\usepackage{listings}
\DeclareCaptionStyle{ruled}{labelfont=normalfont,labelsep=colon,strut=off} 
\floatstyle{ruled}
\newfloat{listing}{tb}{lst}{}
\floatname{listing}{Listing}

\title{Sample-Efficient Behavior Cloning Using General Domain Knowledge}

\author{
Feiyu Zhu
\and
Jean Oh\And
Reid Simmons
\affiliations
Carnegie Mellon University
\emails
\{feiyuz, rsimmons\}@andrew.cmu.edu,
jeanoh@cmu.edu
}

\begin{document}

\maketitle

\begin{abstract}
Behavior cloning has shown success in many sequential decision-making tasks by learning from expert demonstrations, yet they can be very sample inefficient and fail to generalize to unseen scenarios. One approach to these problems is to introduce general domain knowledge, such that the policy can focus on the essential features and may generalize to unseen states by applying that knowledge. Although this knowledge is easy to acquire from the experts, it is hard to be combined with learning from individual examples due to the lack of semantic structure in neural networks and the time-consuming nature of feature engineering. To enable learning from both general knowledge and specific demonstration trajectories, we use a large language model’s coding capability to instantiate a policy structure based on expert domain knowledge expressed in natural language and tune the parameters in the policy with demonstrations. We name this approach the Knowledge Informed Model (KIM) as the structure reflects the semantics of expert knowledge. In our experiments with lunar lander and car racing tasks, our approach learns to solve the tasks with as few as 5 demonstrations and is robust to action noise, outperforming the baseline model without domain knowledge. This indicates that with the help of large language models, we can incorporate domain knowledge into the structure of the policy, increasing sample efficiency for behavior cloning. 
\end{abstract}

\section{Introduction}

Behavior cloning and its variants have demonstrated success in learning policies for autonomous driving \cite{hu2022model}, table-top manipulation \cite{chi2023diffusion}, household tasks \cite{fu2024mobile}, and so on.
Yet, due to a distribution mismatch between expert trajectories and the states encountered during deployment \cite{osa2018algorithmic} as well as the increase in model size, they often rely on a large number of expert demonstrations to learn a robust policy \cite{zhao2024aloha} and cannot generalize well to new camera poses, unseen distractor objects, novel background texture etc. \cite{xie2024decomposing}.

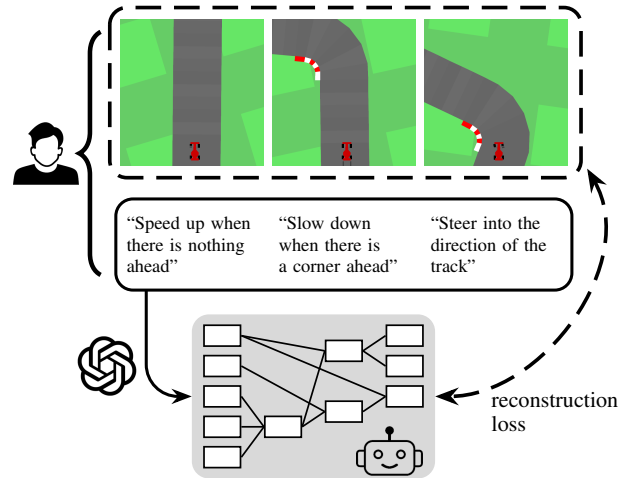
\begin{figure}[t]
    \centering
    \resizebox{0.98\columnwidth}{!}{%
    \begin{tikzpicture}[
        whitenode/.style={rectangle, draw=black, fill=white, ultra thick, minimum width=0.14\columnwidth, minimum height=0.08\columnwidth},
    ]
        \draw[black, rounded corners=16pt, line width=1.0mm] (3.5, -6.5) rectangle (18.5, -9.5);
        \draw[black, rounded corners=16pt, line width=1.2mm, dash pattern=on 24pt off 12pt] (3.3, -0.2) rectangle (18.8, -5.8);
        \node[inner sep=0pt] (expert) at (1.2, -5) {\includegraphics[width=0.25\columnwidth]{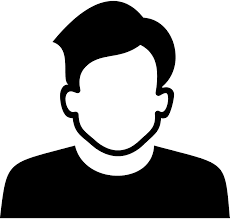}};
        \node[inner sep=0pt] (gpt) at (3.2, -12) {\includegraphics[width=0.25\columnwidth]{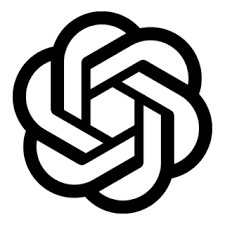}};
        \draw[decorate, decoration={brace, amplitude=20pt, mirror}, line width=1.5mm] (3, -1) -- (3, -9);
        \node[inner sep=0pt] (speed) at (6, -3) {\includegraphics[width=0.55\columnwidth]{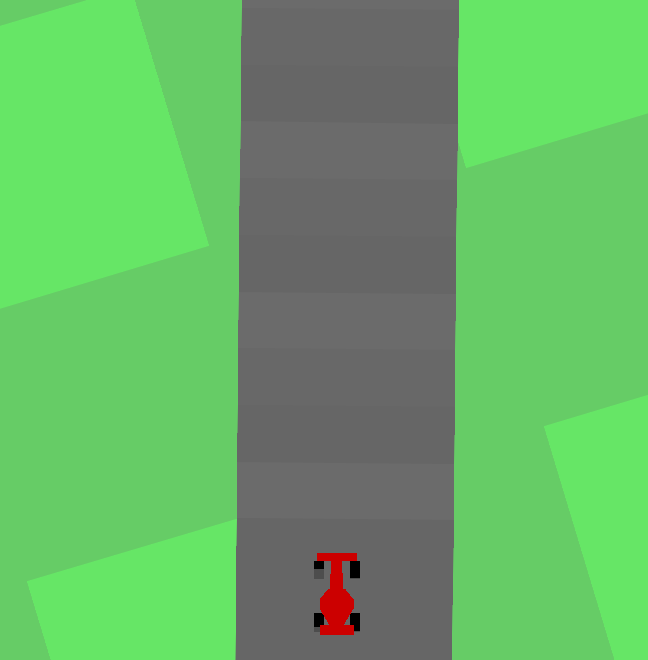}};
        \node[inner sep=0pt] (slow) at (11, -3) {\includegraphics[width=0.55\columnwidth]{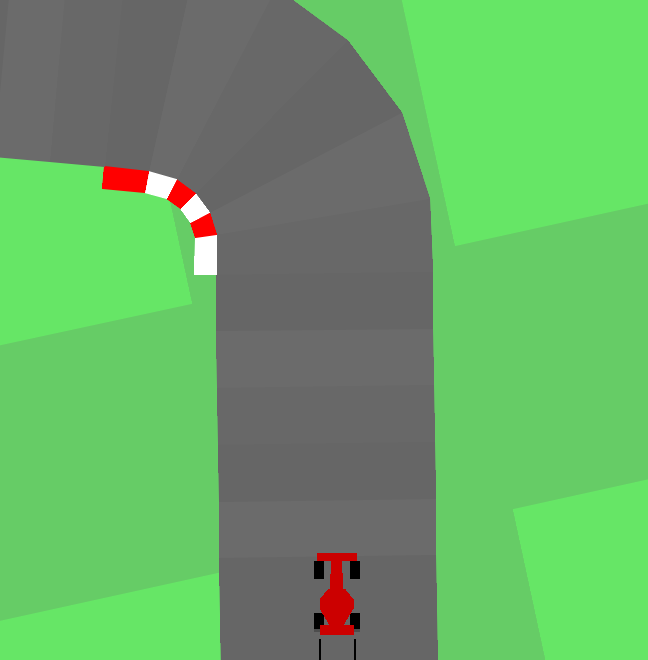}};
        \node[inner sep=0pt] (turn) at (16, -3) {\includegraphics[width=0.55\columnwidth]{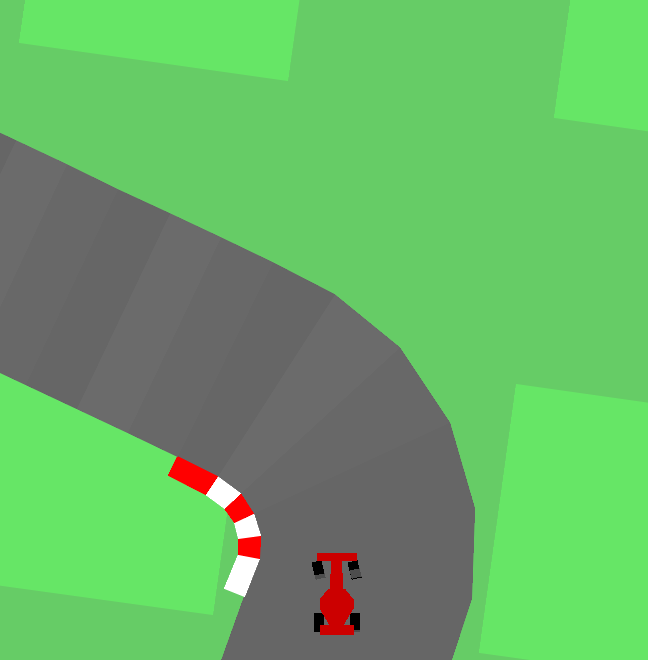}};
        \node[align=left, text width=0.5\columnwidth, font=\huge] at (6, -8) {
            ``Speed up when there is nothing ahead''
        };
        \node[align=left, text width=0.5\columnwidth, font=\huge] at (11, -8) {
            ``Slow down when there is a corner ahead''
        };
        \node[align=left, text width=0.5\columnwidth, font=\huge] at (16, -8) {
            ``Steer into the direction of the track''
        };
        \draw[arrows = {-Stealth[scale=1.2]}, line width=1mm, rounded corners=30pt]  (4.5, -9.5) -- (4.5, -13) -- (6, -13);
        \draw[arrows = {Stealth[scale=1.2]-Stealth[scale=1.2]}, line width=1.2mm, dash pattern=on 24pt off 12pt] (19, -5.5) to[out=-60, in=0] (14, -13);
        
        \fill[white!70!gray, rounded corners=16pt, line width=1.0mm] (6, -10.3) rectangle (14, -15.7);
        \node[inner sep=0pt] (agent) at (12.5, -14.8) {\includegraphics[width=0.25\columnwidth]{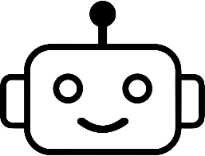}};

        \node[whitenode] (a0) at (7, -11) {};
        \node[whitenode] (a1) at (7, -12) {};
        \node[whitenode] (a2) at (7, -13) {};
        \node[whitenode] (a3) at (7, -14) {};
        \node[whitenode] (a4) at (7, -15) {};
        
        \node[whitenode] (b0) at (9, -14) {};

        \node[whitenode] (c0) at (11, -11.5) {};
        \node[whitenode] (c1) at (11, -13.5) {};
        
        \node[whitenode] (d0) at (13, -11) {};
        \node[whitenode] (d1) at (13, -12) {};
        \node[whitenode] (d2) at (13, -13) {};

        \draw[ultra thick] (a0.east) -- (c0.west);
        \draw[ultra thick] (a0.east) -- (d2.west);
        \draw[ultra thick] (a1.east) -- (c1.west);
        \draw[ultra thick] (a2.east) -- (b0.west);
        \draw[ultra thick] (a3.east) -- (b0.west);
        \draw[ultra thick] (a4.east) -- (b0.west);
        \draw[ultra thick] (b0.east) -- (c0.west);
        \draw[ultra thick] (b0.east) -- (c1.west);
        \draw[ultra thick] (c0.east) -- (d0.west);
        \draw[ultra thick] (c0.east) -- (d1.west);
        \draw[ultra thick] (c1.east) -- (d2.west);
        
        \node[align=left, text width=0.5\columnwidth, font=\Huge] at (18, -13.5) {
            reconstruction loss
        };

    \end{tikzpicture}
    }
    \caption{
        Overview of behavior cloning with general domain knowledge.
        We collect domain knowledge from the expert (middle) in addition to demonstrations (top).
        An LLM translates this knowledge into the structure of the policy (bottom) and behavior cloning is used to learn the parameters of the policy from the demonstrations.
    }
    \label{fig:overview}
\end{figure}

Despite the vast variations a task could have, the underlying principles of solving the task often stay the same.
For example, when trying to open a door, the motion only depends on the position and type of the handle along with the direction the door is expected to open.
Furthermore, the opening direction of the door can be inferred from the location of the hinges.
This general domain knowledge naturally reflects the latent features of the task and their connectivities: \textit{direction} is a latent variable that depends on the location of the hinge, and the specific motion depends on \textit{direction} but not other features such as color or material.
It has been shown that following general knowledge enables zero-shot transfer to novel environments for tasks with discrete action space \cite{zhu2024bootstrapping}.

Although it is relatively easy for a domain expert to explain the general ideas, it is challenging for them to specify the detailed instructions, especially for continuous action spaces.
Additionally, unstructured model architectures have very few semantic structures, creating a barrier between domain knowledge expressed in natural languages and the internal representation of a learning model.
Therefore, existing work that attempts to integrate domain knowledge focuses mainly on state abstractions that highlight the important features of the task \cite{peng2024learning}.

To make use of domain knowledge beyond state representations, we propose \textbf{\textit{Knowledge Informed Models (KIM)}} (Figure \ref{fig:overview}) to take advantage of the coding capabilities of LLMs to instantiate the entire \textit{structure} of the policy, while using expert demonstrations to fit the unspecified parameters in the policy (e.g., how much to slow down when approaching the corner).
This allows the model to tailor not only which input features are used, but also how latent variables should be defined and computed.
This semantically meaningful structure has fewer parameters to be tuned and guides the policy to interpret the demonstrations strategically, so they inherently require fewer samples and are less susceptible to overfitting.

The contributions of this paper are twofold: 1) we propose an approach to make use of general domain knowledge to enable sample-efficient behavior cloning, and 2) we demonstrate the effectiveness and robustness of our approach in continuous environments with discrete and continuous action spaces with very few demonstrations. 
Specifically, our approach achieves better performance than the unstructured baseline with statistical significance and degrades much less than the baseline under the noisy action condition.

\section{Related Work}

\subsection{Sample-Efficient Behavior Cloning}

Data augmentation is a common technique for expanding the coverage of expert demonstrations \cite{ankile2024juicer}, often using visual synthesis \cite{zhou2023nerf}, local continuity \cite{deshpande2024data}, time-reversal symmetry \cite{cheng2024look}.
Extending this direction, other works proposed to learn a local model to guide the policy from unseen states to known states \cite{park2022robust} or to learn a world model \cite{kolev2024efficient}.
Another similar approach is to use state abstraction to hide the irrelevant features of the states \cite{peng2024learning} such that the policy will not be conditioned on them without the need for data augmentation.
Other approaches include using a better representation of actions \cite{chi2023diffusion}, building up skill libraries to reuse previously learned skills \cite{wan2024lotus}, instructing the expert to demonstrate failure recovery \cite{brandfonbrener2023visual}.

Unlike previous work that mainly focused on sample-level operations, our work is the most similar to \cite{mao2023pdsketch} where we aim to improve sample efficiency by specializing the structure of the policy being learned to the specific task and its relevant features.
However, instead of searching through a pre-defined architecture space or merely abstracting the state representations, we take advantage of experts' domain knowledge to instantiate a neural net with a specific structure that is specialized to the task as the policy model.

\subsection{LLM Assisted Policy Learning}

Previous work has used the coding capability of LLMs to implement agent policies \cite{zhu2024bootstrapping}, represent world models \cite{tang2024worldcoder}, generate reward distributions \cite{bucker2024grounding}, and translate underspecified task specifications into structured representations \cite{liu2023llm+}.
However, the codes generated are mostly symbolic, relying on well-defined APIs to execute the policy.
They are also static, allowing very little post-generation adaptation.
Others have used LLMs to generate target action distribution \cite{zhou2024largelanguagemodelpolicy} or provide reward signals \cite{wang2024rl} that can be used to train smaller models.
But unlike the coding-focused approaches that can make use of external knowledge, these sample-based methods depend solely on the pre-trained knowledge in the LLMs.

Our work makes use of the coding ability of LLMs but also enables parameter tuning after code generation.
This alleviates the dependence on the LLMs to get everything correct in one go and makes it possible to incorporate expert knowledge that is not captured by the LLMs.

\subsection{Knowledge Integration in Machine Learning}

It is well-acknowledged that integrating existing human knowledge helps with machine learning \cite{deng2020integrating}, where human knowledge is commonly in the form of feature selections and invariance in the task.

Prior to the popularity of learning feature representations, models were trained with features that were picked manually \cite{bahnsen2016feature} or according to some statistical metrics \cite{ghojogh2019feature}.
Despite achieving great performance in complex tasks such as planning for driving \cite{dauner2023parting}, existing works modify only the inputs to the models but not the architecture of the models, not taking full advantage of the existing domain knowledge.

Other works have developed specialized architecture that incorporates the invariance in the task, including SE(3)-equivariant layers for tabletop manipulation \cite{eisner2024deep} and drug discovery \cite{schneuing2024structure}, and physics-informed neural nets that respect PDE constraints \cite{wang2023expert}.
These approaches require the experts to have both domain knowledge for the task and also engineering skills for model development, and the architecture can only be used in a certain family of tasks.

By contrast, our approach takes advantage of LLMs' coding skills to implement arbitrary general domain knowledge expressed in natural languages, making it more accessible to make use of existing human knowledge. 
And it implements the architecture from the ground up, reflecting both the selection of features and the connections between the features.

\section{Knowledge Informed Model (KIM)}

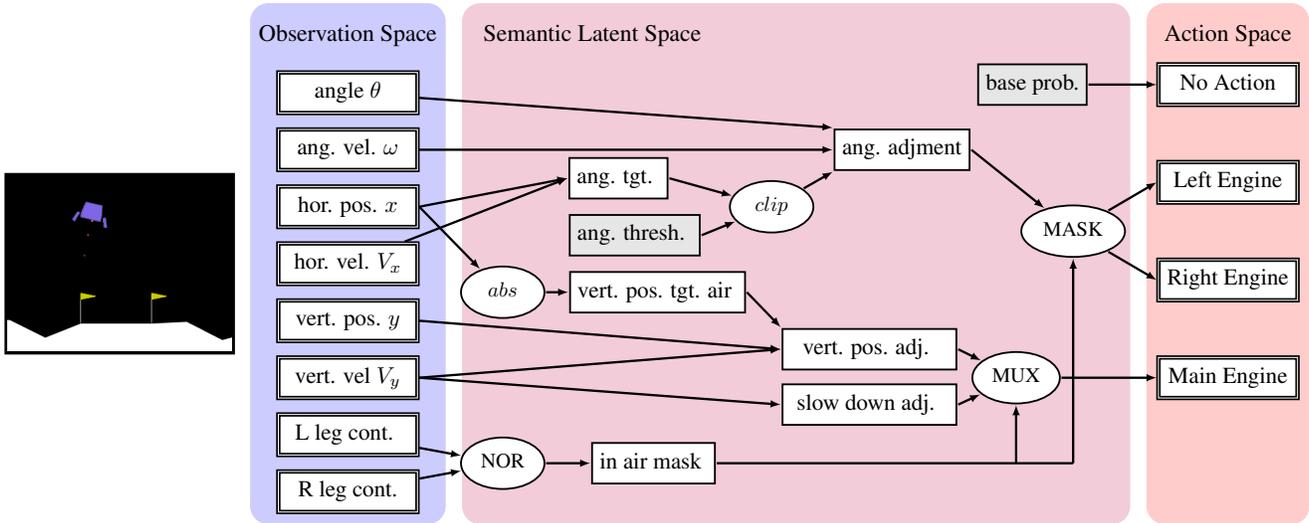
\begin{figure*}[ht]
    \centering
    \resizebox{0.98\linewidth}{!}{
    \begin{tikzpicture}[
        inputnode/.style={rectangle, double, draw=black, fill=white, ultra thick, minimum width=0.12\linewidth, minimum height=0.035\linewidth, scale=2.0, },
        latentnode/.style={rectangle, draw=black, fill=white, ultra thick,  minimum height=0.035\linewidth, scale=2.0, },
        constnode/.style={rectangle, draw=black, fill=white!80!gray, ultra thick,  minimum height=0.035\linewidth, scale=2.0, },
        opnode/.style={shape=ellipse, draw=black, fill=white, ultra thick, scale=1.8, minimum width=0.08\linewidth, minimum height=0.035\linewidth, },
    ]
        \node[inner sep=0pt, draw=black, line width=6pt] (lander) at (3.5, -8) {\includegraphics[width=0.8\columnwidth]{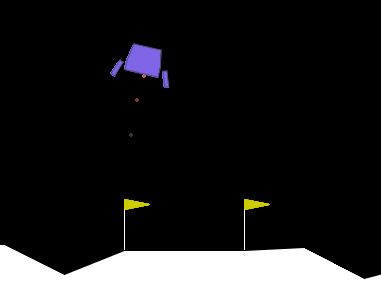}};

        \fill[white!80!blue, rounded corners=16pt, line width=1.0mm] (7.5, 0) rectangle (13.5, -16);
        \fill[white!80!purple, rounded corners=16pt, line width=1.0mm] (14, 0) rectangle (34.5, -16);
        \fill[white!80!red, rounded corners=16pt, line width=1.0mm] (35, 0) rectangle (40, -16);
        \node[font=\Huge] at (10.5, -1) { Observation Space };
        \node[font=\Huge] at (18, -1) { Semantic Latent Space };
        \node[font=\Huge] at (37.5, -1) { Action Space };

        \node[inputnode] (theta) at (10.5, -2.75) { angle $\theta$\strut };
        \node[inputnode] (omega) at (10.5, -4.5) { ang. vel. $\omega$\strut };
        \node[inputnode] (hpos) at (10.5, -6.25) { hor. pos. $x$\strut };
        \node[inputnode] (hvel) at (10.5, -8) { hor. vel. $V_x$\strut };
        \node[inputnode] (vpos) at (10.5, -9.75) { vert. pos. $y$\strut };
        \node[inputnode] (vvel) at (10.5, -11.5) { vert. vel $V_y$\strut };
        \node[inputnode] (L) at (10.5, -13.25) { L leg cont. \strut };
        \node[inputnode] (R) at (10.5, -15) { R leg cont.\strut };

        \node[latentnode] (angtgt) at (18.8, -5.375) { ang. tgt. \strut };
        \node[constnode] (angthr) at (19.3, -7.125) { ang. thresh. \strut };
        \node[latentnode] (verttgt) at (20, -8.875) { vert. pos. tgt. air \strut };
        \node[latentnode] (inair) at (19.875, -14.125) { in air mask \strut };

        \node[opnode] (logair) at (15.25, -14.125) { NOR\strut };
        \node[opnode] (abs) at (15.25, -8.875) {$abs$\strut};
        \node[opnode] (clip) at (23.5, -6.25) {$clip$\strut};
        \node[opnode] (logori) at (32.75, -7) { MASK\strut };
        \node[opnode] (select) at (31, -11.5) { MUX\strut};
        
        \node[latentnode, minimum width=0.15\linewidth,] (vertadj) at (26.5, -10.625) { vert. pos. adj. \strut };
        \node[latentnode, minimum width=0.15\linewidth,] (sdadj) at (26.5, -12.325) { slow down adj. \strut };

        \node[latentnode] (angadj) at (27.5, -4.5) { ang. adjment\strut };

        \node[constnode] (base) at (31.5, -2.5) { base prob.\strut};
        
        \node[inputnode] (NOOP) at (37.5, -2.5) { No Action\strut };
        \node[inputnode] (LE) at (37.5, -5.5) { Left Engine\strut };
        \node[inputnode] (RE) at (37.5, -8.5) { Right Engine\strut };
        \node[inputnode] (ME) at (37.5, -11.5) { Main Engine\strut };

        \draw[-latex, line width=2pt] (base) -- (NOOP);
        \draw[-latex, line width=2pt] (theta) -- (angadj.north west);
        \draw[-latex, line width=2pt] (omega) -- (angadj);
        \draw[-latex, line width=2pt] (hpos.east) -- (angtgt.west);
        \draw[-latex, line width=2pt] (hvel) -- (angtgt.west);
        \draw[-latex, line width=2pt] (angtgt.east) -- (clip);
        \draw[-latex, line width=2pt] (angthr.east) -- (clip);
        \draw[-latex, line width=2pt] (clip) -- (angadj.south west);
        \draw[-latex, line width=2pt] (angadj.east) -- (logori);
        \draw[-latex, line width=2pt] (logori) -- (LE.west);
        \draw[-latex, line width=2pt] (logori) -- (RE.west);
        \draw[-latex, line width=2pt] (hpos.east) -- (abs);
        \draw[-latex, line width=2pt] (abs) -- (verttgt);
        \draw[-latex, line width=2pt] (verttgt.east) -- (vertadj.north west);
        \draw[-latex, line width=2pt] (vpos.east) -- (vertadj.west);
        \draw[-latex, line width=2pt] (vvel.east) -- (vertadj.west);
        \draw[-latex, line width=2pt] (vertadj.east) -- (select);
        \draw[-latex, line width=2pt] (vvel.east) -- (sdadj.west);
        \draw[-latex, line width=2pt] (sdadj.east) -- (select);
        \draw[-latex, line width=2pt] (select) -- (ME);
        \draw[-latex, line width=2pt] (L) -- (logair);
        \draw[-latex, line width=2pt] (R) -- (logair);
        \draw[-latex, line width=2pt] (logair) -- (inair);
        \draw[-, line width=2pt] (inair) -- (32.75, -14.125);
        \draw[-latex, line width=2pt] (32.75, -14.125) -- (logori);
        \draw[-latex, line width=2pt] (31, -14.125) -- (select);
        
    \end{tikzpicture}
    }
    \caption{
    Illustration of the Knowledge Informed Model for the Lunar Lander environment generated by GPT.
    Each box represents a variable, with the white boxes representing latent variables and the gray boxes representing tunable parameters in the model.
    Arrows represent the dependencies between variables and each has an associated learnable weight.
    The oval shapes represent non-linear operations.
    By default the value of latent variables is a linear combination of the variables that it depends on.
    }
    \label{fig:lander_model}
\end{figure*}

%
%
%
%

\subsection{Structured Policy}

In this work, we use the term ``structured policy" to refer to a model in which latent variables and their connectivities are specialized to the task.
The latent variables typically have semantic meanings, representing key features of the task that are not directly accessible from the input.
A structured policy has many distinctions compared to an unstructured model such as a generic multi-layer perception (MLP).
It takes advantage of the sparsity that exists in many domains \cite{mao2023pdsketch} and assigns learnable parameters to the related latent variables instead of all pairs of latent variables.
It may also contain a variety of operations (e.g., \texttt{max}, \texttt{clip}) that are beyond linear transformations and primitive non-linear activations.
As a result, the policy structure is a highly concise representation of the general structure of a solution to the task (e.g., Figure \ref{fig:lander_model}).
Mathematically, a policy structure can be represented as a 4-tuple $\langle V, O, E, \Theta\rangle$ where
\begin{itemize}[noitemsep, topsep=0pt]
    \item $V$ is a set of nodes that each represent a latent variable.
    \item $O$ is a set of nodes that each represent an instance of an operation (e.g., the \texttt{clip} function).
    \item $E: \{ \langle u_i, v_i \rangle \}$  is a set of edges that represents the dependencies between latent variables and operations.
    \item $\Theta: E \to \mathbb{R}$ is a set of weights associated for each edge.
\end{itemize}
\noindent In general, it can be seen as a weighted acyclic graph.

During inference time, latent variables are computed in an order based on the partial order of their dependencies.
That is, the latent variables that only depend on the input features are computed first (1\textsuperscript{st} degree latent), then the variables that only depend on the input features and the 1\textsuperscript{st} degree latent, and so on.
The specific values of the variables depend on the operations defined by the policy structure and the weight parameters connecting the latent variables.
For instance, as shown in Figure \ref{fig:lander_model}, the value of angle adjustment is a linear combination (with bias) of the current angle, current angular velocity, and the angle target after clipping.

The weights in $\Theta$ can be updated via gradient descent by supervised learning on the action output, similar to how a typical MLP can be learned.
This enables numeric learning in structured policies.
Section \ref{sec:bc} provides a detailed explanation of the learning process.

Although a structured policy along with its parameter values can be directly coded by an expert (e.g., the heuristics-based policy in the Lunar Lander task \cite{towers2024gymnasiumstandardinterfacereinforcement}), doing so manually is typically time-consuming.
Therefore, to enable more scalability, it would be beneficial that the structure of the policy be generated from natural language descriptions and the parameters be learned from a few demonstrations, which are easier to acquire from an expert.

\subsection{KIM Generation via LLM}

We assume access to domain knowledge $\mathcal{K}$ from an expert in natural language that describes the high-level ideas that guide the demonstrations $\mathcal{D}$.
This is attainable as previous works have shown that humans typically construct simplified mental representations during problem-solving \cite{ho2022people}, so they should also be able to articulate the general knowledge used to perform the demonstrations.

In practice, we collect the general description of the strategy used by the expert, the feature space and the action space of the task, and any additional information about the environment that might be useful.
We can instantiate a policy structure based on the provided general knowledge using an LLM (e.g., GPT4o \cite{openai2024gpt4ocard}).
That is
\begin{align}
   \langle V, O, E, \Theta_\text{init} \rangle = LLM(\mathcal{S} + \mathcal{K})
\end{align}
\noindent where $\mathcal{S}$ is the system prompt that is shared for all tasks.

Concretely, we use the chain-of-thought prompting \cite{wei2023chainofthoughtpromptingelicitsreasoning} to instruct the LLM to implement the models.
First, it is instructed to extract all the input features and latent variables in the general knowledge description and list their type and shape (e.g., angle target has type \texttt{float} and shape \texttt{(1,)}).

Next, the LLM is expected to re-arrange the latent variables in the order in which they should be computed based on variable dependencies (e.g., \texttt{angle target} should appear before \texttt{angle adjustment} as the latter depends on the former). 
And, for each of them, list all the previously computed variables that the current variable depends on and what operators are needed to connect them.
Empirical experiments showed that without this step the LLM may miss some of the connections in the code generation process.
During this process, the LLM is also instructed to classify each connection between latent variables as ``positively correlated" or ``negatively correlated" based on the expert knowledge (e.g., ``the target angle depends on the horizontal position and should point to the center" indicates that the target angle and the current horizontal position are positively correlated).
This information can be used to set the initial value of the parameters $\Theta_\text{init}$.
Since the general knowledge does not contain specific numeric relationships, we let the LLM set very rough initial values (e.g., $-0.1$ for negatively correlated variables).
As the model structure reflects the semantic meaning of the expert's strategy, it does not have the permutation symmetry as many unstructured models do \cite{ainsworth2022git} and hence is more sensitive to the initial values.

The final step is to implement the structure as a subclass of \texttt{nn.Module} in PyTorch.
When coding the model, the LLM is instructed to classify all the parameters as gradient or non-gradient, where the non-gradient parameters are those that cannot be learned using gradient descent (e.g., the bounds of a \texttt{clip} function) whereas the rest are gradient parameters.

Figure \ref{fig:lander_model} shows an example of a model generated for the Lunar Lander task.
The prompts that were used to generate this model can be found in the Appendix.
Note that we do not include any examples in the prompt, and fully leverage the zero-shot coding capability of the LLM.

\subsection{Behavior Cloning for KIM}\label{sec:bc}

After the model structure is set, we train the parameters using the standard behavior cloning objective to tune the parameters $\Theta$ in the policy.
\begin{align}
    \min_\Theta \mathbb{E}_{\langle s_i, a_i \rangle \in \mathcal{D}}\mathcal{L}(a_i, \pi_\Theta(s_i))
\end{align}
In practice, we use grid search over the values for the non-gradient parameters. 
For each combination of the non-gradient parameter values, we use gradient descent to optimize the remaining gradient parameters.
By default, we use cross-entropy loss for discrete action spaces and mean square error for continuous action spaces.
The combination (both non-gradient and gradient) that achieves the least overall loss is kept as the final model parameter.

Because the connections between latent variables are sparse, the number of total parameters is small compared to unstructured models.
Additionally, we focus on using only a few demonstrations.
Therefore, we can perform gradient descent on all the demonstrations at once for most tasks without having to separate the samples into mini-batches.
This helps to stabilize the training process.

Unlike unstructured models, latent variables in KIM have semantic meanings as they are extracted from the provided expert knowledge.
Therefore, it is possible for the expert to directly set the value of some constant parameters, or the weights connecting different latent variables.
This will make learning easier since there are fewer parameters to optimize.

\section{Experiments}

We experiment with the Lunar Lander and Car Racing environments implemented in Gymnasium \cite{towers2024gymnasiumstandardinterfacereinforcement}.
The environments cover both discrete and continuous action spaces.
We used \texttt{gpt-4o-2024-11-20} as our LLM for all of the experiments.

\subsection{Lunar Lander}

The objective of the Lunar Lander task is to control the engines of the lander to perform a soft landing on the landing pad (an illustration can be found in Figure \ref{fig:lander_model}).
The observation space is the horizontal position and velocity, vertical position and velocity, angular position and velocity, and whether each of the landing legs is in contact with the surface.
The last two features on the landing legs are binary, while the others are continuous.
Each new episode has a different randomly initialized starting configuration.
The action space is discrete, consisting of doing nothing or activating one of the left, main, or right engines.
The episode ends if the lander lands safely, crashes, or runs out of fuel after $1000$ steps.
Typically, a successful landing can be achieved in around $200$ steps.

We use the heuristic policy defined in the Gymnasium package as the expert policy that generates demonstrations.
This policy achieves a $90\%$ success rate in the environment, however
we keep only the successful episodes as demonstrations for training.
We manually describe the strategy of the heuristic policy as the expert general knowledge and use it to prompt the LLM for KIM generation.
The prompt can be found in the Appendix.

For the baseline condition, we use an MLP and formulate it as a classification problem with cross-entropy loss, where the objective is to predict which action the expert policy is going to perform given all of the features of a state.
For both conditions, we randomly sample $20\%$ of the demonstration steps as the validation set, and keep the model parameters with the least loss in the validation set for evaluation.

\subsection{Car Racing}

\begin{figure}[t]
    \centering
    \includegraphics[width=\columnwidth]{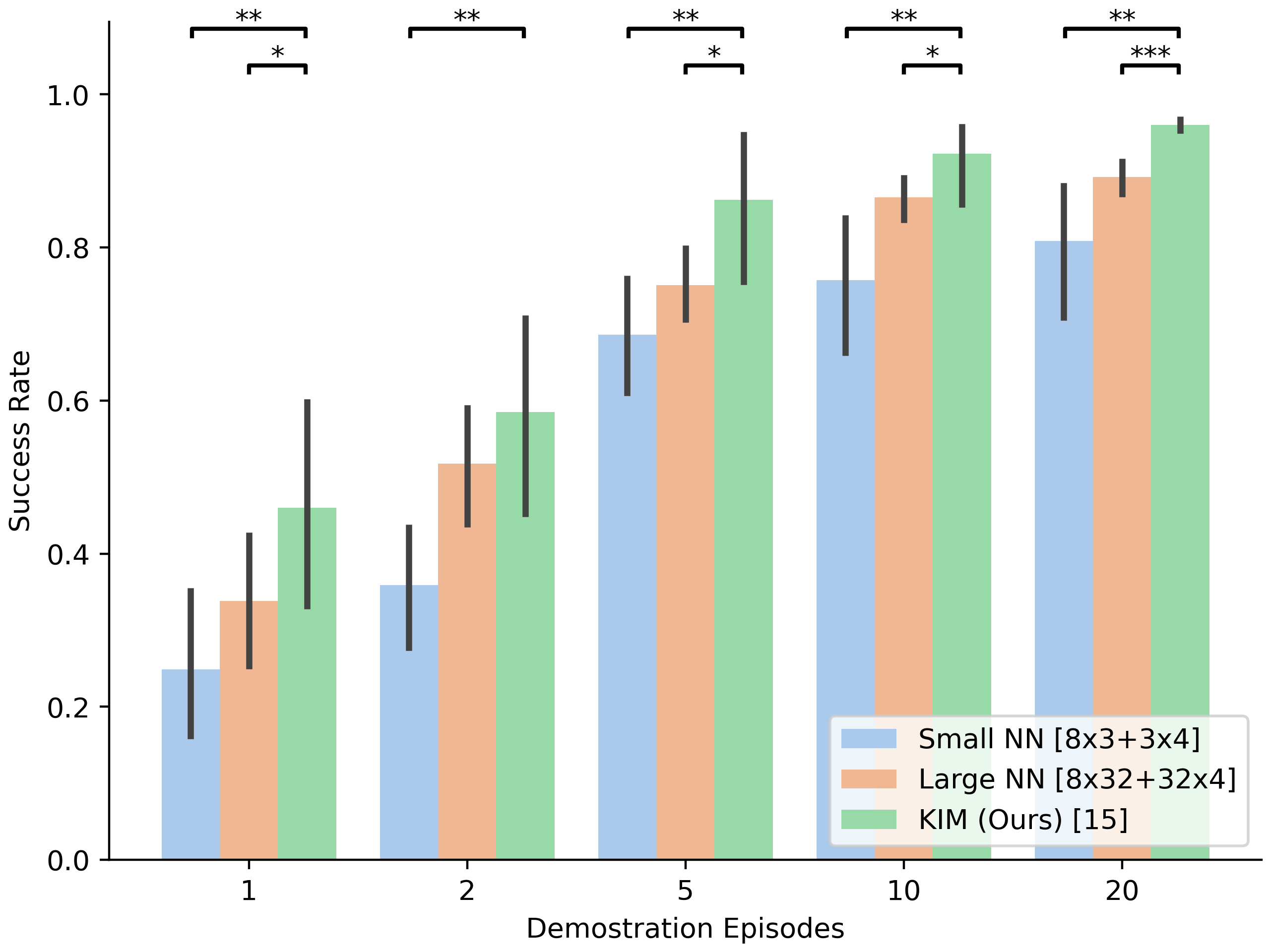}
    \caption{
    Success rates in the Lunar Lander task, evaluated on $100$ random start states per session.
    The error bars in the plot show the $95\%$ confidence interval estimated by $20$ sets of demonstration episodes.
    Asterisks denote the statistical significance levels of paired t-tests (* for $< 0.05$, ** for $< 0.01$, and *** for $< 0.001$).
    }
    \label{fig:lander_success}
\end{figure}

The objective of the car racing task is to complete a winding track as fast as possible (an illustration is provided in Figure \ref{fig:overview}).
The track is defined by a sequence of tiles that span from the left of the track to the right.
The reward is defined as 
\begin{align}
    1000 * N - 0.1 * T
\end{align}
where $N$ is the percentage of the tiles on the track covered within $1000$ steps and the $T$ is the number of time steps taken to complete the track (or truncated at $1000$ if the race car runs out of time).
A tile is covered if at least one of the wheels makes contact with it.
In addition to the first $1000$ steps, after recording the reward for the environment, we keep running the environment for another $2000$ steps to collect the maximum coverage of the track for a policy.
Empirically this is sufficient to wait for the policy to finish the track at least once.

The original environment features an image-based observation space.
To bypass the perception challenges, we use a basic representation where each state is defined by a sequence of tiles (their mid-point coordinates, angular heading, the difference in coordinates and headings compared to the previous tile, whether the tile has a corner marker) that makes up the visible tracks in the current frame and the current state of the race car (including current speed, direction or steer, value of the gyroscope, and ABS sensors on each of the wheels).
Each new episode features a new track layout.
The action space is continuous and consists of the steering of the race car, the engagement of the gas pedal, and the engagement of the brake pedal.
The gas pedal only acts on the rear wheels while the brakes are on all wheels.
The exact dynamics of the race car are unavailable to the human expert and the model.

In this environment, the domain knowledge and demonstration trajectories all come from a human researcher who has interacted with the environment extensively.
Specifically, the actions are collected using a Logitech controller for continuous actions while the researcher is looking at the rendering shown on a screen.
A total of $200$ demonstration episodes are collected through multiple sessions.
This setup reflects the real-world scenario where the same expert provides the domain knowledge while giving demonstrations that correspond to the domain knowledge.
Since there are human errors during execution and discrepancy in the perception modalities (i.e., the policy takes in low dimensional input while the human expert operates on images), there will be noise in the demonstration provided, further resembling real-world settings.
All demonstrations achieve perfect coverage (i.e., the race car never goes off the track), and have an average reward of $913.5$.

The baseline condition is an MLP that takes the same basic representation as input.
Its learning objective is to minimize the mean squared error between its output and the expert action in the demonstration set.
Similar to the Lunar Lander environment, $20\%$ of the expert demonstrations are reserved for the validation set while the rest are used for training in both conditions.

\subsection{Results on Learning with a Few Demonstrations}

\begin{figure}[ht]
    \centering
    \includegraphics[width=\columnwidth]{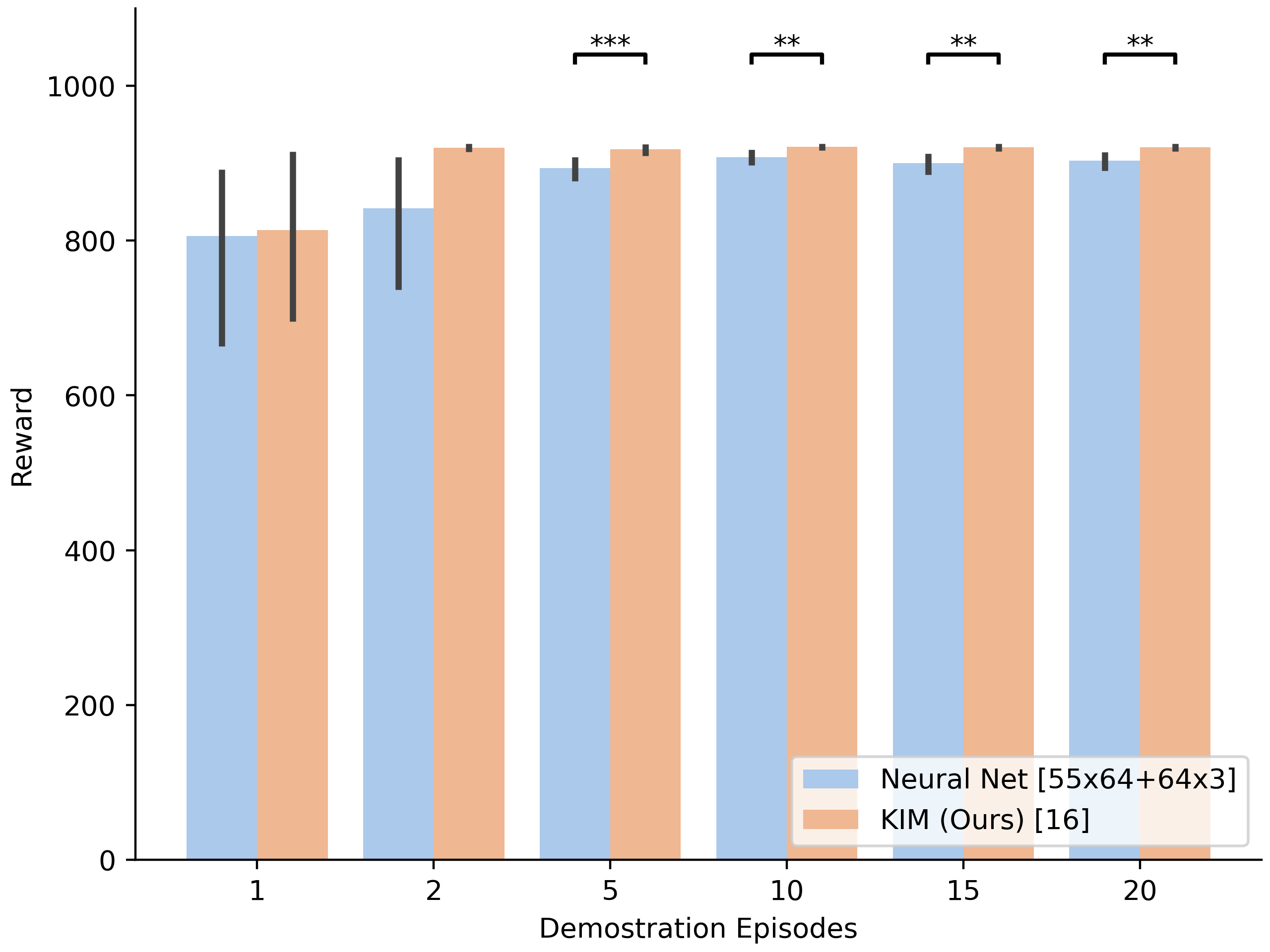}
    \caption{
    Reward in the Car Racing task, evaluated on $100$ random tracks per session.
    The error bars in the plot show the $95\%$ confidence interval estimated by $10$ sets of demonstration episodes.
    }
    \label{fig:racing_reward}
\end{figure}


\begin{figure}[ht]
    \centering
    \includegraphics[width=\columnwidth]{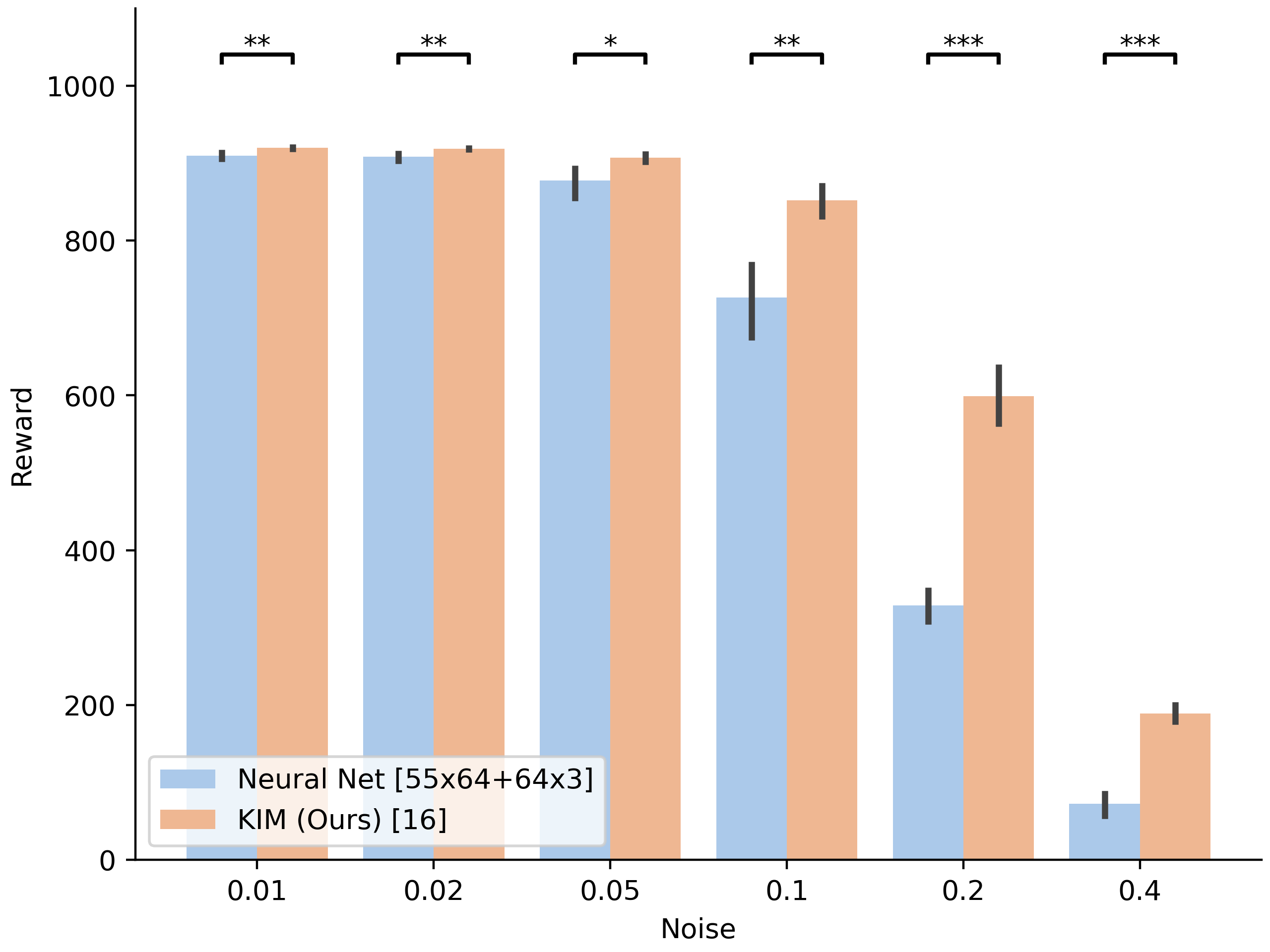}
    \caption{
    Reward in the Car Racing task with different levels of action noise, each evaluated on $100$ random tracks per session.
    The error bars in the plot show the $95\%$ confidence interval estimated by $10$ sets of demonstration episodes.
    Each model is trained with $10$ demonstration episodes.
    }
    \label{fig:racing_noisy}
\end{figure}

\begin{figure}[ht]
    \centering
    \resizebox{\linewidth}{!}{
    \begin{tikzpicture}
        \node[inner sep=0pt] (img) at (0, 0) {
            \includegraphics[width=\columnwidth]{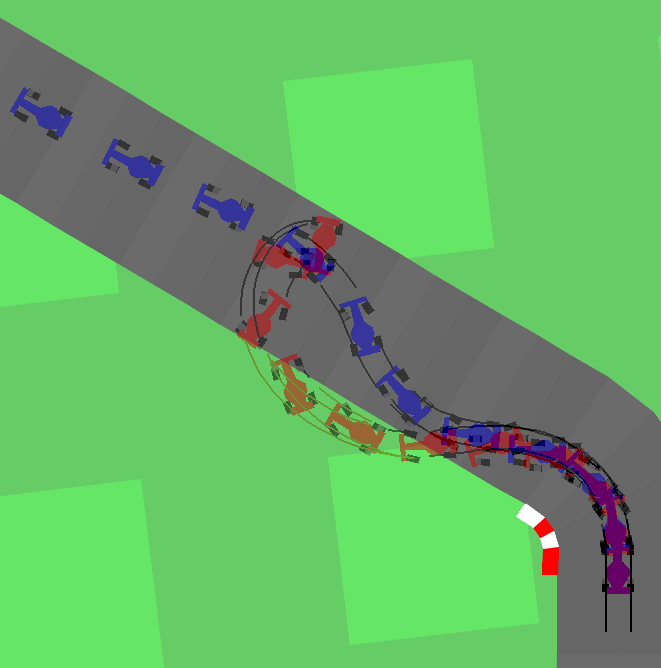}
        };
        \fill[white, rounded corners=2pt, line width=1.0mm] (1.4, 2.4) rectangle (4, 4);
        \node[inner sep=0pt, rotate=90] (bcar) at (2.0, 2.8) {
            \includegraphics[width=0.05\columnwidth]{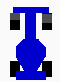}
        };
        \node[inner sep=0pt, rotate=90] (rcar) at (2.0, 3.6) {
            \includegraphics[width=0.05\columnwidth]{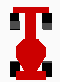}
        };
        \node[align=left, text width=0.3\columnwidth, font=\small] at (3.7, 2.8) {
            KIM (Our)
        };
        \node[align=left, text width=0.3\columnwidth, font=\small] at (3.7, 3.6) {
            Neural Net
        };
    \end{tikzpicture}
    }
    \caption{
    Qualitative samples from the neural net baseline (red) and KIM (blue) in the same starting condition. The two models are trained on the same set of $2$ demonstrations. The positions of the race car are captured with a fixed time interval in between. The baseline over-steers and loses control while KIM corrects the heading of the race car.
    }
    \label{fig:racing_sample}
\end{figure}
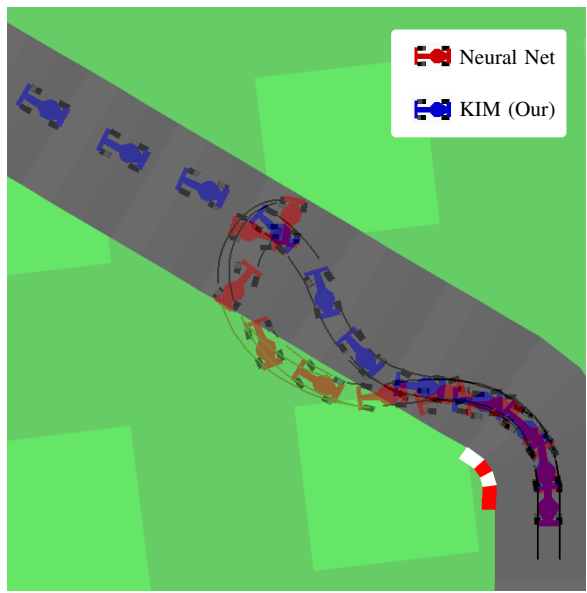

\begin{figure}[ht]
    \centering
    \resizebox{\linewidth}{!}{
    \begin{tikzpicture}
        \node[inner sep=0pt] (img) at (0, 0) {
            \includegraphics[width=\columnwidth]{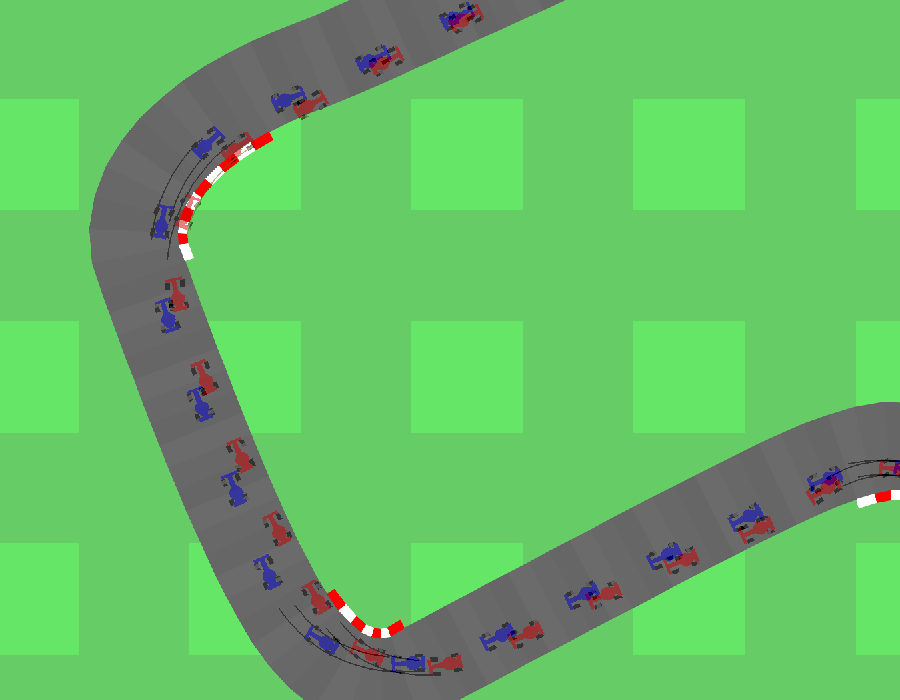}
        };
        \fill[white, rounded corners=2pt, line width=1.0mm] (1.4, 1.4) rectangle (4, 3);
        \draw[yellow, rounded corners=2pt, line width=1.0mm] (-1.3, 0.7) rectangle (-3.5, 2.5);
        \node[inner sep=0pt, rotate=90] (bcar) at (2.0, 1.8) {
            \includegraphics[width=0.05\columnwidth]{figures/icons/bluecar.png}
        };
        \node[inner sep=0pt, rotate=90] (rcar) at (2.0, 2.6) {
            \includegraphics[width=0.05\columnwidth]{figures/icons/redcar.png}
        };
        \node[align=left, text width=0.3\columnwidth, font=\small] at (3.7, 1.8) {
            KIM (Our)
        };
        \node[align=left, text width=0.3\columnwidth, font=\small] at (3.7, 2.6) {
            Neural Net
        };
    \end{tikzpicture}
    }
    \caption{
    Stability comparison between the two conditions.
    Both models are trained on the same set of $10$ demonstrations.
    The baseline takes the corner very tightly (it is on the curb in the yellow box) while KIM is closer to the center line.
    Despite traveling for a longer distance, KIM takes less time to navigate through the two corners.
    }
    \label{fig:racing_center}
\end{figure}


Figure \ref{fig:lander_success} shows the success rate in the Lunar Lander task between using KIM and the baseline neural networks of two different sizes (the number of parameters are listed in the legend).
The figure shows that given a fixed number of demonstrations, KIM, with only $15$ parameters, achieves a $20\%+$ higher success rate than the small NN that has a similar number of parameters and a $7\%+$ higher success rate than the larger NN that has $25$x more parameters.
This shows that given the same demonstrations, KIM learns a more robust policy.
All but one pair of comparisons show statistical significance in a paired t-test where the pairing is based on having the same set of demonstrations.
The plot on the rewards of the Lunar Lander environment can be found in the Appendix.

Figure \ref{fig:racing_reward} shows the reward comparisons between KIM and a neural net in the Car Racing task.
It shows that starting from as few as $2$ demonstrations KIM yields good performance and low variance.
This attributes to KIM having very few parameters organized in a semantically meaningful structure and is thus more robust to imperfect demonstrations.
When there are more demonstrations, KIM still outperforms the baseline (which has $200$x more parameters) with statistical significance.
Overall, the plots show that despite the demonstrations and the provided general knowledge may not be perfectly aligned, using the knowledge to instantiate the model still leads to better performance.

\subsection{Result on Environments with Noise}

To evaluate how well KIM does in noisy settings, we randomly corrupt its output with a Gaussian noise when it is interacting with the environment. 
That is,
\begin{align}
    \mathbf{a}_\mathrm{new} \sim \mathcal{N}(\mathbf{a}_\mathrm{pred};~ \mathrm{noise\_level} \cdot \mathbf{I})
\end{align}
Note that in this setting the models are still trained with expert demonstrations captured in a noise-free environment.
The noise is only added after the model is trained. 

Figure \ref{fig:racing_noisy} shows the comparison between KIM and the baseline (using the $10$ models trained on $10$ demonstrations each in the previous section) in environments with different noise levels.
It shows that as the noise level increases, the performance of the baseline condition degrades much more drastically than KIM.
In particular, KIM can still retain around $65\%$ of the reward even with considerable noise (the action space is $[-1, 1]$ and the Gaussian noise has a standard deviation of $0.2$).
This illustrates that optimizing with respect to general domain knowledge makes the policy less brittle.

\subsection{Qualitative Analysis}

\begin{listing}[tb]%
\caption{Code snippet generated by GPT on steering control}%
\label{lst:code}%
\begin{lstlisting}[language=Python]
steer_control = (
    self.steer_weight * 
    (target_heading - current_heading) * 
    (1 - current_speed)
)
\end{lstlisting}
\end{listing}

The high variance in the neural net baseline learned with $2$ demonstrations is partially caused by the learned model losing control of the race car and swirling off the track.

Figure \ref{fig:racing_sample} shows an example where the baseline condition loses control while KIM steers the race car to stay on the track.
This is because the knowledge of ``don't steer too drastically when accelerating" is crucial for driving the rear-wheel drive race car in this domain, yet is only implicitly illustrated through the expert demonstrations.
As a result, when the demonstrations are not sufficiently indicative of this, an unstructured model may miss this constraint, leading to catastrophic outcomes, especially if no expert demonstration illustrates how to regain control after losing traction.

However, KIM provides another channel (i.e., the general knowledge in natural language) for the expert to pass knowledge to the learning model.
Furthermore, these domain constraints are enforced by the structure of the model such that it is more robust to imperfect demonstrations.
Therefore it can navigate the corner much more smoothly.
Listing \ref{lst:code} shows the structure that is informed by the following instructions:

\begin{displayquote}
the output of steering should be scaled based on the current speed such that when speed approaches $1$ the steer magnitude should approach $0$.
\end{displayquote}
The full prompt and model can be found in the Appendix.

We observe a similar case where the race car driven by KIM follows the center line more closely, following the expert's instructions, while the baseline neural network takes the corners more tightly (Figure \ref{fig:racing_center}).
This helps explain why KIM is more robust to action noise, since it is less likely to leave the track.

\subsection{Additional Comparisons}

\begin{table*}[t]
    \centering
    \begin{tabular}{l|ccc}
        \toprule
         & Lunar Lander Success Rate $\uparrow$ & Car Racing Reward $\uparrow$ & Car Racing Coverage $\uparrow$ \\
        \midrule
        KIM & $0.891~(\pm 0.184)$ & $921.631~(\pm 10.251)$ & $1.000~(\pm 0.002) $ \\
        KIM w/ human-generated code & $0.880~(\pm 0.086)$ & $922.048~(\pm ~~7.892) $ & $1.0~(\pm 0.0)$\\
        KIM w/ random parameters & $0.702~(\pm 0.390)$ & $824.371~(\pm 79.916)$ & $0.999~(\pm 0.005)$\\
        \midrule
        Expert policy & $0.890$ & $913.498~(\pm 10.86)$ & $1.0~(\pm 0.0)$ \\
        \bottomrule
    \end{tabular}
    \caption{Comparison of KIM in different settings. KIMs are trained with $10$ demonstrations.
    For the Lunar Lander environment, the success rate is computed among $100$ randomly initialized configurations. 
    We learn $10$ models each trained with a different set of demonstrations to estimate the mean and standard deviation of success rates.
    For the Car Racing environment, the mean and standard deviation are evaluated on $100$ randomly initialized track layouts for a single model. 
    }
    \label{tab:ablation}
\end{table*}

Table \ref{tab:ablation} shows the ablation on different settings of KIM in the two environments.

The human-generated code condition is where the code generated by GPT is replaced with code generated by a human researcher given the same prompt.
Overall, the codes generated are very similar and hence the performance is similar in both tasks.
The differences in the code implementation are very subtle.
For example, the human-generated code set \texttt{bias=False} for the linear layer for adjusting for steering because by default the race car should go straight.
However, GPT did not make use of this information and defined the linear layer with bias.
Details like this likely lead to a slightly smaller variance in the human-generated code condition.

On the other hand, there is a distinctive difference between having pre-filled initialization values for the parameters and not.
In the random initialization condition, all parameters are sampled from a standard normal distribution.
The result shows very high variance, confirming that the optimization space for KIM does not have the ``one basin" phenomenon \cite{ainsworth2022git} that helps optimize typical unstructured models.
So prompting the LLM to analyze the relationship between latent variables is vital for performance.

Additionally, KIM performed slightly better than the expert demonstration in the Car Racing environment.
This is likely due to the imperfections in the human demonstrations and KIM's ability to filter those imperfections in the demonstrations based on general knowledge, which is hard for an unstructured model as it treats all demonstrations equally.

\section{Discussion}

\subsection{Limitations}

The current method relies heavily on having access to good expert instructions.
These are relatively easy to acquire from well-established settings (e.g., assembly lines or aircraft controls) but could be hard in scenarios that require more nuance (e.g., social navigation).
It is also challenging for human experts to provide exhaustive instructions in one go, or if the action space is different between human experts and the policy (e.g., learning a quadruped robot walking policy).

Another limitation is the dependency on LLMs' zero-shot coding capabilities.
In addition to the potential misalignment issue \cite{greenblatt2024alignment}, all contemporary LLMs operate on input sequences with a length limit, making it impossible to generate a KIM if the general domain knowledge exceeds that limit.
Additionally, previous work has reported that LLMs may neglect information in a long prompt \cite{liu2024lost}, which may lead to a suboptimal structure.

\subsection{Future Work}

Since it is hard to specify all the general knowledge all at once, a natural extension is to support interactive and incremental KIM.
This would require integrating the code repairing capability of LLM \cite{tang2024coderepairllmsgives} and modifying the existing structure based on the incoming knowledge.
The parameters are expected to transfer to the new architecture with a little fine-tuning.
This also helps if the LLM does not generate the correct code the first time, by giving the human expert opportunities to amend the generated model.

Additionally, the benefit of integrating general knowledge applies beyond representing policies - it can also be used to represent the transition model in the world.
One could use a similar technique to develop a sample-efficient model-based reinforcement learning policy where the structure of the world comes from some existing database and the specific parameters are tuned by interacting with the real world.

Finally, although current experiments have shown that vanilla gradient descent works on KIM with some non-linearity, more work needs to be done to investigate how well this approach scales to more complex domains where the connections in the structure are more complex.

\section{Conclusion}

In this work, we proposed Knowledge Informed Models (KIM) that combine expert demonstrations with general domain knowledge by instantiating a policy structure from the general knowledge before tuning its parameters with expert demonstrations.
This bridges the gap between the semantic knowledge human experts typically possess and the unstructured model architectures that are used for behavior cloning.
We detailed how an LLM can be used to enable structure generation and how it could be learned from gradient descent once the initial values of the parameters are set.
Through the Lunar Lander and Car Racing tasks, we show that our approach is more sample-efficient than an unstructured baseline and also more robust to noisy environments after training.
We have also presented qualitatively how having a structure enables more robustness to imperfect expert demonstrations.
Finally, we discussed the limitations of this work and how it can be extended in the near future.

\clearpage
\bibliographystyle{named}
\bibliography{ijcai25}

\appendix
\onecolumn
\section*{Technical Appendix}

\section{More Results for Learning with a Few Demonstrations}

\begin{figure}[h]
    \centering
    \begin{subfigure}[t]{0.45\textwidth}
        \centering
        \includegraphics[width=\textwidth]{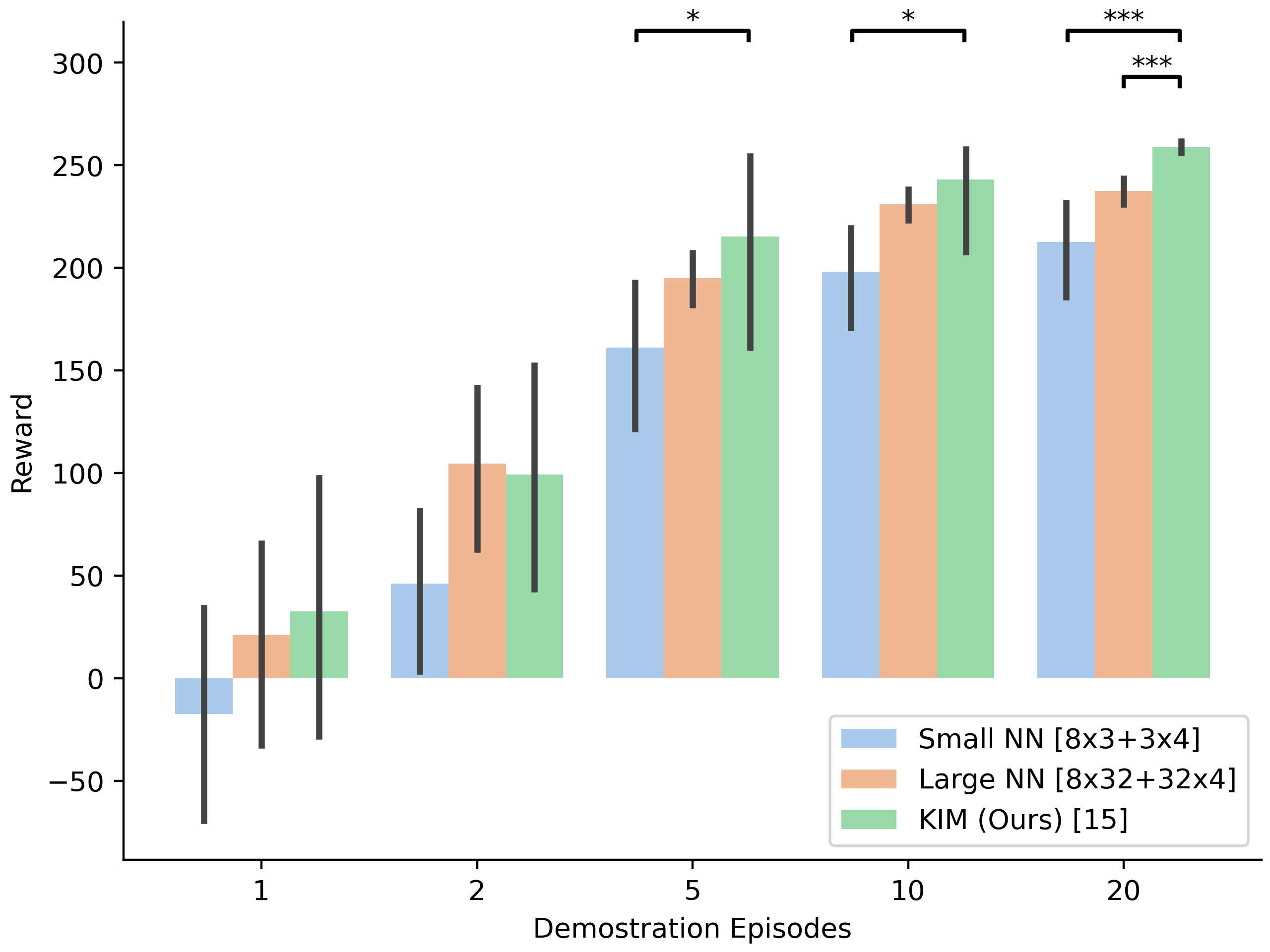}
        \caption{
        Rewards in the Lunar Lander environment, evaluated on $100$ random start states per session.
        The error bars in the plot show the $95\%$ confidence interval estimated by $20$ sets of demonstration episodes.
        Asterisks denote the statistical significance levels of paired t-tests (* for $< 0.05$, ** for $< 0.01$, and *** for $< 0.001$).
        }
        \label{fig:lander_reward}
    \end{subfigure}
    \hfill
    \begin{subfigure}[t]{0.45\textwidth}
        \centering
        \includegraphics[width=\textwidth]{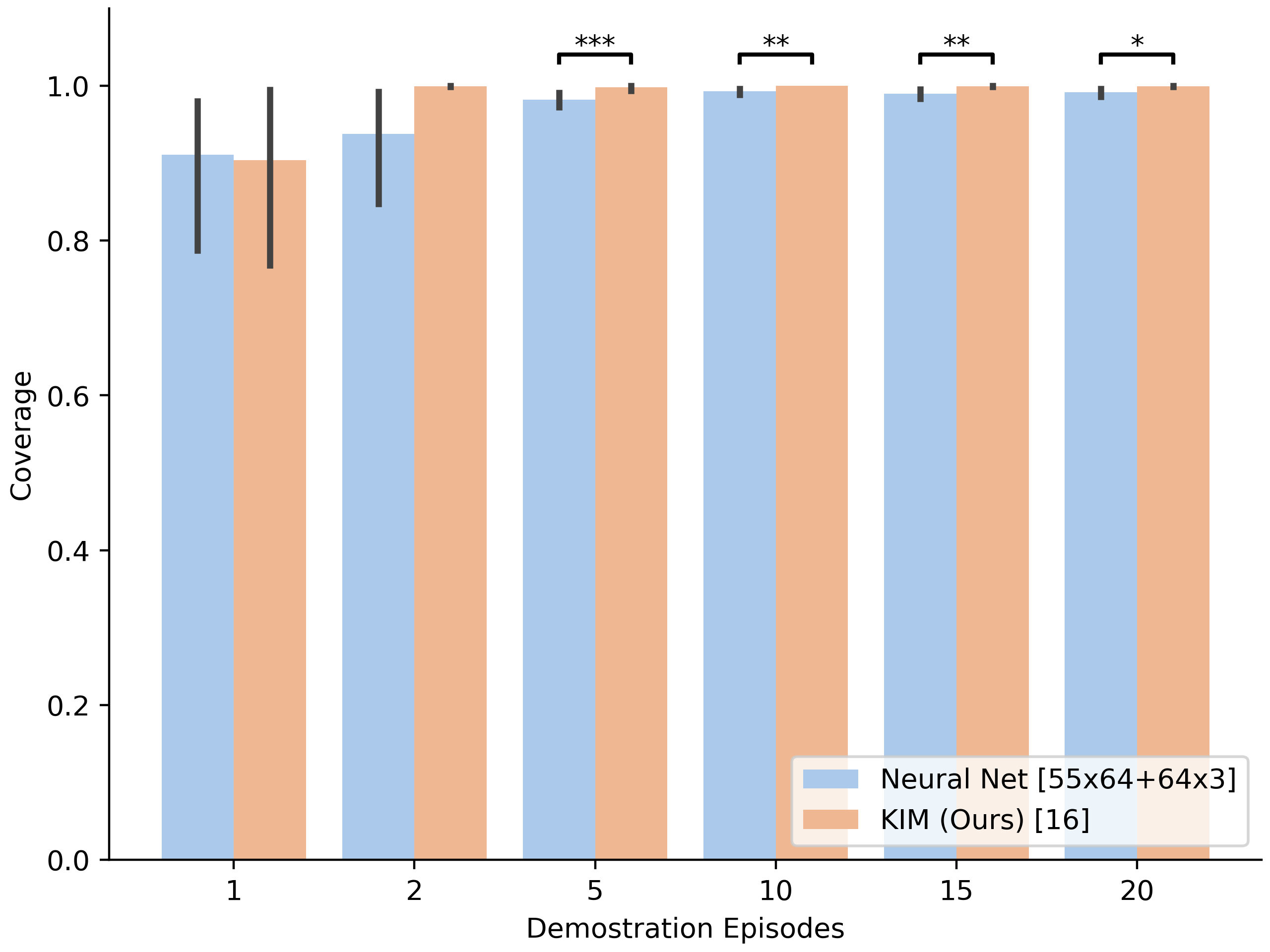}
        \caption{Coverage in the Car Racing task, evaluated on $100$ random tracks per session.
        The error bars in the plot show the $95\%$ confidence interval estimated by $10$ sets of demonstration episodes.}
        \label{fig:racecar_coverage}
    \end{subfigure}
    \caption{More results on the Lunar Lander task and the Car Racing task}
    \label{fig:extra_res}
\end{figure}

Figure \ref{fig:extra_res} shows the reward in the Lunar Lander task and the coverage in the Car Racing task when learning from a few demonstrations.
Overall they show a very similar trend to the plots included in the main text.

The specific reward definition for the Lunar Lander is not explicitly disclosed in the Gymnasium descriptions, but is correlated to the distance to the landing pad, the current speed of the lander, the tilt of the lander, whether the engines are activated, whether the lander has crashed or landed successfully.
This is not as indicative as the success rate for the Lunar Lander environment as the expert policy does not specifically optimize for reward, and there should be no expectation that the behavior learning model should achieve a high reward.


The main reason for not achieving perfect coverage is typically due to taking the shortcut through corners, such that not all tiles on the corners are visited.
In some cases, for instance, in the qualitative example presented in the main content, the race car loses control and spins off the track into an unrecoverable state.

\section{More Qualitative Examples of Car Racing}

\begin{figure}[h]
    \centering
    \begin{subfigure}[t]{0.45\textwidth}
        \centering
        \includegraphics[width=\textwidth]{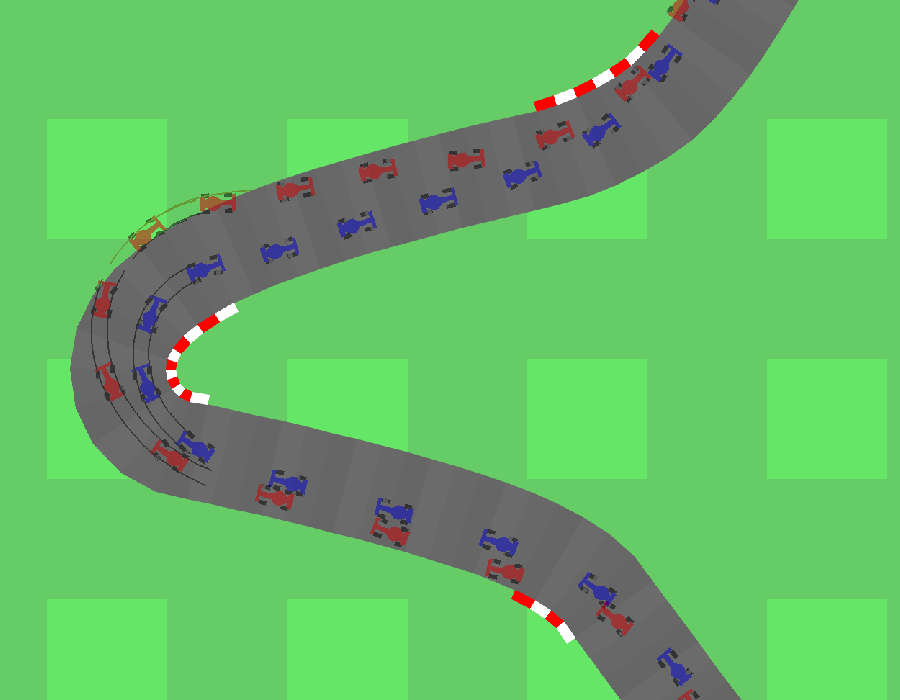}
        \caption{
        }
        \label{fig:stacked3}
    \end{subfigure}
    \hfill
    \begin{subfigure}[t]{0.45\textwidth}
        \centering
        \includegraphics[width=\textwidth]{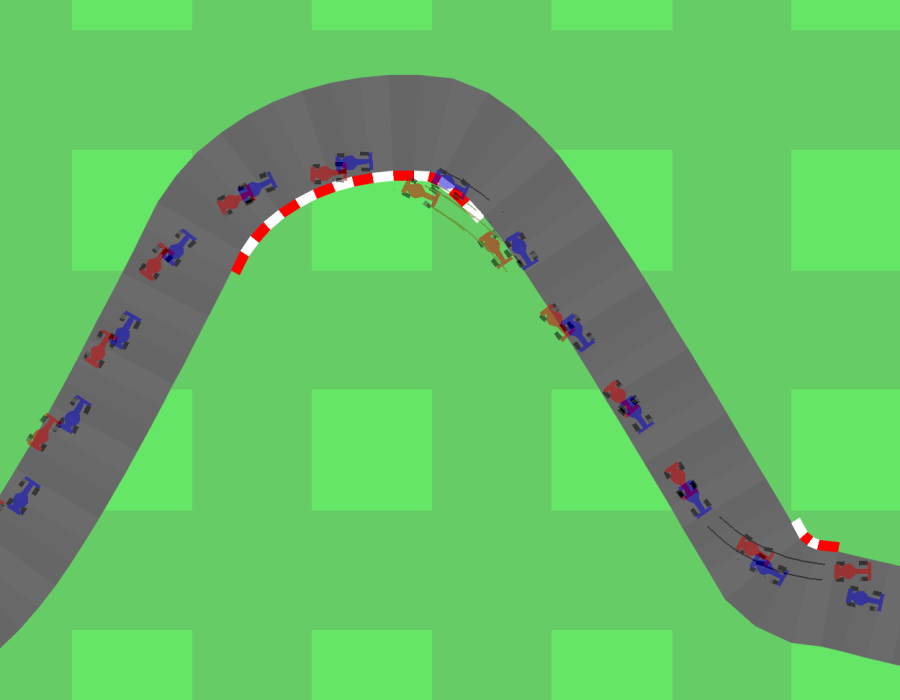}
        \caption{}
        \label{fig:stacked4}
    \end{subfigure}
    \caption{More samples from the neural net (red) and KIM (blue) rollouts. Both models are trained on the same set of $10$ demonstrations.}
    \label{fig:extra_rollout}
\end{figure}

Figure \ref{fig:extra_rollout} shows more samples of the baseline policy and KIM.
It is shown that KIM policy is more stable during corning.
By contrast, sometimes the neural net policy takes the turn too loose (Figure \ref{fig:stacked3}) while sometimes it takes the turn too tight (Figure \ref{fig:stacked4}).
All of these samples show qualitatively why KIM has lower variance and is more robust to action noise.

\section{GPT Prompts}

The following is used as the \textit{system prompt}:

\begin{Verbatim}[breaklines, fontsize=\small, numbers=right]
Implement pytorch models that follow the specified structure of the user.

The user will provide the following:
 * [Structure Description] explains the connections / operations between each variables / features
 * [Features] explains how to interpret the input to the model. For example, which dimension of the input corresponds to which feature, and their type (discrete or continuous)
 * [Output Space] explains the action space
 * [Additional Notes] (optional) explains any additional details of the task

You will do the following steps before giving the final implementation:
 * [Variables] Extract and list the intermediate (latent) variable (if any) from the user's description. Also, indicate the shape or type of each variable.
 * [Plan the connections] List the variables (and their type and shape) in the order in which they should be computed, where the variables based only on the input features are listed first, then the variables that depend on those, etc. For each of them, explain how they can be computed using the previously listed variables or inputs to the model. Also indicate whether the new feature is positively correlated to the previous feature or negatively correlated respectively. Be very specific. List the functions or operators that should be used to connect the variables. If you decided to use a linear combination, explain why a bias term is included or not included.
 * [Code] Provide the implementation of the model as a subclass of `nn.Module`. Do not include examples. No need to explain the code.

Use the following format:
[Variables]
 * Name of the first variable (shape and type)
 * Name of the second variable (shape and type)
...

[Connections]
 * Name of the first variable that should be computed (shape and type)
    - depends on <feature 1 name> (positively correlated), <feature 2 name> (nagatively correlated), ...
    - can be computed with a linear combination of ... and with a bias term
    - the bias term is included because ...
 * Name of the second variable that should be computed (shape and type)
    - depends on <feature 1 name> (negatively correlated), <feature 2 name> (positively correlated), ...
    - can be computed using `torch.where` on ...

[Code]
```py
import torch
from torch import nn

class ModelName(nn.Module):
    <YOUR MODEL DEFINITION>
```


Notes:
 * Represent new features as linear combinations of old features when possible. If you are very certain no bias term is needed (i.e., the feature value should be 0 if all inputs are 0), then don't include the bias term to make it easier to learn. Register all weights and bias terms as `nn.Parameter` with `required_grad=True`.
 * There might be cases where linear combination is not sufficient, then you may use other operations such as multiplication to represent the interaction between two features.
 * When initializing the values for each weight and bias, set a value based on positive or negative correlation (e.g., if the input feature is negatively correlated then set its weight to -0.1) instead of using random initialization.
 * There should be no constant number in the `forward` function. If there is a parameter that can't be learned by gradient descent (e.g., the bounds in the clamp function), then label and register it as a `nn.Parameter` in the model class. Also, use the comment to label it as a "non-gradient parameter".
 * You may use any of the functions defined in pytorch (e.g., `torch.logical_and`, `torch.clamp`, `torch.abs`, `torch.square`, etc.).
 * Make sure the gradient can flow back to the parameters. Avoid in-place operations (use new variable names instead) or constructing new tensors (use `torch.stack` or `torch.cat` instead).
 * You may assume the inputs are already normalized.
 * For discrete output space, the model should output a (potentially unnormalized) distribution among those discrete actions. For continuous action space, return the predicted action without worrying about distributions, but clip the values to the appropriate range (if specified).
 * At any point in the process, if you are unsure about something, or if there is some ambiguity, then state so and ask a clarification question instead of proceeding.
\end{Verbatim}

The following is the prompt used for the Lunar Lander task

\begin{Verbatim}[breaklines, fontsize=\small, numbers=right]
[Structure Description]
The lander is in air if none of its legs are in contact with the ground. Otherwise, it is in contact with the ground.

The target heading of the lander depends on its horizontal coordinate and speed so that it points to the center. But we will clip in a range such that it stays roughly in the middle, because tilting too much is bad.

The target vertical coordinate depends on the magnitude of the horizontal offset of the lander. The further the lander is to the landing pad (which is at $(0, 0)$), the higher the target vertical coordinate should be.

The heading adjustment depends on the difference between the current and clipped target heading of the lander as well as the current angular velocity.

The speed adjustment needed to put the lander to rest is proportional to its vertical speed.

And the vertical adjustment depends on the difference between the current and target vertical coordinate as well as the vertical speed.

Only activate the left or the right engine when the lander is not contacting the ground. And the probability of activating the left engine is the heading adjustment, and symmetrically the probability of activating the right engine is the negation of the heading adjustment.

The probability of activating the main engine in air is the vertical adjustment.

The probability of activating the main engine when the lander is in contact with the ground is the speed adjustment.

There is a base level probability that the lander will do nothing regardless of the input.

[Features]
The input to the model is a tensor of $(8)$. The features of the lander in each dimension are:
0. (float32) horizontal coordinate $x$
1. (float32) vertical coordinate $y$
2. (float32) horizontal speed $v_x$
3. (float32) vertical speed $v_y$
4. (float32) heading $\theta$
5. (float32) angular velocity $\omega$
6. (bool) whether the left landing leg is in contact with the ground
7. (bool) whether the right landing leg is in contact with the ground

[Output Space]
A tensor of shape (4,) representing the unnormalized distribution among the following four actions
0: do nothing
1: fire left orientation engine
2: fire main engine
3: fire right orientation engine

[Additional Notes]
The lander is upright when $\theta = 0$ and is tilting to the left when $\theta > 0$. When the lander is falling $v_y < 0$ since the y-axis points upward.
\end{Verbatim}

The following is the prompt used for the Car Racing task.

\begin{Verbatim}[breaklines, fontsize=\small, numbers=right]
[Structure Description]
First, find the tile that is close to the race car using the tile's xy coordinates (the race car is at the origin). Then the tiles close to the race car are defined as the consecutive tiles that follow the closest tile up until some number threshold. Similarly, the tiles ahead of the race car are also the consecutive tiles that follow the closest tile up until some larger threshold or the end of the sequence.

The curvature of two consecutive tiles is defined as the difference between the angle of the tiles (normalized to (-1, 1)) and will be provided. It is signed with a negative value corresponding to the track bending to the left, and a positive value means bending to the right. There is a corner if the absolute value of the curvature is greater than some threshold. The sharpness of the corner is the magnitude of the curvature (e.g., -0.99 corresponds to a sharp left turn).

The target speed when there are no corners ahead can be represented as a constant.

The target speed when there is at least one corner tile ahead is the minimum of the corresponding target speed of each tile. It depends on the magnitude of its lateral distance to the race car, the magnitude of the heading difference, longitudinal distance, and curvature. Specifically, the larger the lateral distance is, the faster the target speed can be because the car doesn't need to make a sharp turn. The larger the longitudinal distance is, the faster the target speed is because the car has more time to slow down later. However, the sharper the corner the lower the target speed should be. And the larger the heading difference is the lower the speed should be because the race car has to steer more.

The target heading of the race car is related to the average heading and curvature of the tiles close to the race car as well as the average lateral position of the race car relative to those tiles. If the lateral position is negative then the race car's target heading is proportionally positive to re-center the race car, and vice versa. But the target heading should also follow the heading and the curvature of the close tiles.

The race car is controlled by three dimensions: steer, accelerate, and brake. Each of them should be controlled by a linear controller that takes in the difference between the target and the current state of the race car's heading and speed and outputs a control signal. However, the output of steering should be scaled based on the current speed such that when speed approaches $1$ the steer magnitude should approach $0$. That is, when the current speed is high the steer magnitude should be discounted so that the race car does not lose control, and when the speed is low it can steer with more magnitude.

[Features]
The input to the model is two tensors representing the tracks and other indicator information:
0. (float32, (L, 8)) tiles. Where $L$ is the number of tiles on the track and the last dimension contains $(x, y, _, _, \theta, _, \delta_\theta, _)$ where $x$ is the signed lateral position of the tile, $y$ is the signed longitudinal position of the tile, $\theta$ is the heading of the tile, and $\delta_\theta$ is the different in heading between neighboring tiles (curvature). Other features from input (marked by $_$) can be ignored.
1. (float32, (7,)) indicators. Where the first column is the speed $v$, and the second column is the current heading of the race car $\theta$ while the other features are not important.

[Output Space]
A tensor of shape (3,) where each element corresponds to steer, accelerate, and brake controls of the race car respectively. The steer value should be in the range of $(-1, 1)$, while the accelerate and brake should be in the range of $(0, 1)$.

[Additional Notes]
The track is a loop.
\end{Verbatim}

\section{Models generated by GPT}

The following is the code generated by GPT for the Lunar Lander task (formatted by an automatic formatter Black\cite{black2020}).

\begin{lstlisting}[language=Python]
import torch
from torch import nn


class LanderModel(nn.Module):
    def __init__(self):
        super().__init__()
        # Learnable parameters for linear combinations
        self.target_heading_weights = nn.Parameter(
            torch.tensor([0.1, 0.1]), requires_grad=True
        )
        self.target_heading_bias = nn.Parameter(torch.tensor(0.0), requires_grad=True)
        self.heading_adjustment_weights = nn.Parameter(
            torch.tensor([-0.1, 0.1, -0.1]), requires_grad=True
        )
        self.heading_adjustment_bias = nn.Parameter(
            torch.tensor(0.0), requires_grad=True
        )
        self.speed_adjustment_weight = nn.Parameter(
            torch.tensor(-0.1), requires_grad=True
        )
        self.vertical_adjustment_weights = nn.Parameter(
            torch.tensor([-0.1, 0.1, -0.1]), requires_grad=True
        )
        self.vertical_adjustment_bias = nn.Parameter(
            torch.tensor(0.0), requires_grad=True
        )
        self.prob_nothing = nn.Parameter(torch.tensor(0.1), requires_grad=True)
        # Non-gradient parameters
        self.target_heading_clip = nn.Parameter(
            torch.tensor([-0.5, 0.5]), requires_grad=False
        )

    def forward(self, x):
        # Extract features
        (
            x_coord,
            y_coord,
            v_x,
            v_y,
            theta,
            omega,
            left_leg_contact,
            right_leg_contact,
        ) = x.unbind(-1)

        # Compute in_air
        in_air = torch.logical_not(
            torch.logical_or(left_leg_contact, right_leg_contact)
        ).float()

        # Compute target_heading
        target_heading = torch.clamp(
            x_coord * self.target_heading_weights[0]
            + v_x * self.target_heading_weights[1]
            + self.target_heading_bias,
            self.target_heading_clip[0],
            self.target_heading_clip[1],
        )

        # Compute target_vertical
        target_vertical = torch.abs(x_coord)

        # Compute heading_adjustment
        heading_adjustment = (
            theta * self.heading_adjustment_weights[0]
            + target_heading * self.heading_adjustment_weights[1]
            + omega * self.heading_adjustment_weights[2]
            + self.heading_adjustment_bias
        )

        # Compute speed_adjustment
        speed_adjustment = v_y * self.speed_adjustment_weight

        # Compute vertical_adjustment
        vertical_adjustment = (
            y_coord * self.vertical_adjustment_weights[0]
            + target_vertical * self.vertical_adjustment_weights[1]
            + v_y * self.vertical_adjustment_weights[2]
            + self.vertical_adjustment_bias
        )

        # Compute probabilities
        prob_left_engine = heading_adjustment * in_air
        prob_main_engine = vertical_adjustment * in_air + speed_adjustment * (
            1 - in_air
        )
        prob_right_engine = -heading_adjustment * in_air

        # Stack probabilities
        probs = torch.stack(
            [self.prob_nothing, prob_left_engine, prob_main_engine, prob_right_engine],
            dim=-1,
        )

        return probs
\end{lstlisting}

The following is the code generated by GPT for the Car Racing task

\begin{lstlisting}[language=Python]
import torch
from torch import nn


class RaceCarController(nn.Module):
    def __init__(
        self, num_close_tiles, num_ahead_tiles, corner_threshold, target_speed_no_corner
    ):
        super().__init__()
        self.num_close_tiles = num_close_tiles  # non-gradient parameter
        self.num_ahead_tiles = num_ahead_tiles  # non-gradient parameter
        self.corner_threshold = nn.Parameter(
            torch.tensor(corner_threshold), requires_grad=False
        )
        self.target_speed_no_corner = nn.Parameter(
            torch.tensor(target_speed_no_corner), requires_grad=False
        )

        # Parameters for target speed with corner
        self.lateral_weight = nn.Parameter(torch.tensor(0.1))
        self.longitudinal_weight = nn.Parameter(torch.tensor(0.1))
        self.heading_weight = nn.Parameter(torch.tensor(-0.1))
        self.curvature_weight = nn.Parameter(torch.tensor(-0.1))
        self.target_speed_bias = nn.Parameter(torch.tensor(0.5))

        # Parameters for target heading
        self.heading_weight_close = nn.Parameter(torch.tensor(0.1))
        self.curvature_weight_close = nn.Parameter(torch.tensor(0.1))
        self.lateral_weight_close = nn.Parameter(torch.tensor(-0.1))
        self.target_heading_bias = nn.Parameter(torch.tensor(0.0))

        # Parameters for control signals
        self.steer_weight = nn.Parameter(torch.tensor(0.1))
        self.accelerate_weight = nn.Parameter(torch.tensor(0.1))
        self.brake_weight = nn.Parameter(torch.tensor(0.1))

    def forward(self, tiles, indicators):
        # Closest tile index
        distances = torch.sqrt(tiles[:, 0] ** 2 + tiles[:, 1] ** 2)
        closest_tile_index = torch.argmin(distances)

        # Close and ahead tiles
        close_tiles = tiles[
            closest_tile_index : closest_tile_index + self.num_close_tiles
        ]
        ahead_tiles = tiles[
            closest_tile_index : closest_tile_index + self.num_ahead_tiles
        ]

        # Corner mask and sharpness
        corner_mask = torch.abs(ahead_tiles[:, 6]) > self.corner_threshold
        sharpness = torch.abs(ahead_tiles[:, 6])

        # Target speed with corner
        target_speed_with_corner = torch.min(
            self.lateral_weight * torch.abs(ahead_tiles[:, 0])
            + self.longitudinal_weight * torch.abs(ahead_tiles[:, 1])
            + self.heading_weight * torch.abs(ahead_tiles[:, 4])
            + self.curvature_weight * sharpness
            + self.target_speed_bias
        )

        # Target speed
        target_speed = torch.where(
            corner_mask.any(), target_speed_with_corner, self.target_speed_no_corner
        )

        # Target heading
        average_heading_close = close_tiles[:, 4].mean()
        average_curvature_close = close_tiles[:, 6].mean()
        average_lateral_close = close_tiles[:, 0].mean()
        target_heading = (
            self.heading_weight_close * average_heading_close
            + self.curvature_weight_close * average_curvature_close
            + self.lateral_weight_close * average_lateral_close
            + self.target_heading_bias
        )

        # Control signals
        steer_control = (
            self.steer_weight * (target_heading - indicators[1]) * (1 - indicators[0])
        )
        accelerate_control = self.accelerate_weight * (target_speed - indicators[0])
        brake_control = self.brake_weight * (indicators[0] - target_speed)

        # Clamp outputs
        steer_control = torch.clamp(steer_control, -1, 1)
        accelerate_control = torch.clamp(accelerate_control, 0, 1)
        brake_control = torch.clamp(brake_control, 0, 1)

        return torch.stack([steer_control, accelerate_control, brake_control])    
\end{lstlisting}

\section{Training Details}

\begin{table*}[h]
    \centering
    \begin{tabular}{c|cc}
        \toprule
         & Lunar Lander & Car Racing \\
        \midrule
        KIM & $n=4000$ @ $lr=0.03$ & $n=200$ @ $lr=0.03$ \\
        Neural Net Baseline & $n=10000$ @ $lr=0.001$ & $n=250$ @ $lr=0.003$ \\
        \bottomrule
    \end{tabular}
    \caption{Number of optimization steps and learning rate for each condition in each task}
    \label{tab:lr}
\end{table*}

We use the Adam optimizer \cite{kingma2017adammethodstochasticoptimization} to learn the model parameters.
All hyperparameters are kept as default in PyTorch except for the learning rate.
The learning rate and number of optimization steps are listed in Table \ref{tab:lr}.
During preliminary experiments we experiment with the learning rate in the range of $0.001$ to $0.05$ and use the largest learning rate that still ensures a converging loss curve.
The number of steps are determined by finding the convergence on the loss curve.

In all conditions, in each step, the gradients are computed based on all demonstrations as one single batch.
No sampling or shuffling is needed.
For the Lunar Lander task we found that no normalization is needed as the values have small magnitude (all in the range of $-4$ to $4$).
But for Car Racing task the coordinates are in pixel values and can range from $0$ to $1000$ while some other information like angles are only in the range of $-\pi$ to $\pi$. 
Therefore, we normalize all values to be between $[-1, 1]$ for both the baseline and KIM in the Car Racing task.

A weight-balanced cross-entropy loss is used for the Lunar Lander task for both conditions.
Mean squared error loss is used for the Car Racing task for both conditions.

When prompting the GPT, we set the seed to $0$ and temperature to $0$.
But we found that even doing so, the output is still not deterministic.

\end{document}